\documentclass[10pt,twocolumn,letterpaper]{article}

\usepackage[pagenumbers]{cvpr} 

\usepackage[accsupp]{axessibility} 

\usepackage{graphicx}
\usepackage{amsmath}
\usepackage{amssymb}
\usepackage{booktabs}
\usepackage{color, colortbl}
\usepackage[dvipsnames]{xcolor}
\usepackage[utf8]{inputenc} 
\usepackage[T1]{fontenc}    
\usepackage{url}            
\usepackage{amsfonts}       
\usepackage{nicefrac}       
\usepackage{microtype}      
\usepackage{multirow}   
\usepackage{algorithm}
\usepackage{algorithmic}
\usepackage{wrapfig}
\usepackage{comment}
\usepackage{pifont}
\usepackage[multiple]{footmisc}
\usepackage{makecell}
\usepackage{caption}
\usepackage{adjustbox}
\usepackage{diagbox} 
\usepackage{changepage}
\usepackage{makecell}
\usepackage{subcaption}
\usepackage{arydshln}
\usepackage{symbols}

\usepackage{mathalfa}
\usepackage{graphicx}
\usepackage{graphbox}
\usepackage{array}
\usepackage{enumitem}
\usepackage{animate}
\usepackage{xspace} 

\AtBeginDocument{%
 \abovedisplayskip=6pt plus 3pt minus 5pt
 \abovedisplayshortskip=0pt plus 3pt
 \belowdisplayskip=6pt plus 3pt minus 5pt
 \belowdisplayshortskip=5pt plus 3pt minus 4pt
}

\makeatletter
\renewcommand{\paragraph}{%
  \@startsection{paragraph}{4}%
  {\z@}{0.4ex \@plus 1ex \@minus .1ex}{-1em}%
  {\normalfont\normalsize\bfseries}%
}
\makeatother

\makeatletter
\@namedef{ver@everyshi.sty}{} %
\makeatother

\newlength{\itemheight} %
\usepackage{tikz} %
\usetikzlibrary{calc} %
\usetikzlibrary{tikzmark} %
\usetikzlibrary{spy} %
\usetikzlibrary{shapes.misc} %
\usepackage{pgfplots} %
\usepackage{pgfplotstable} %
\pgfplotsset{compat=newest} %

\definecolor{darkred}{rgb}{0.7,0.1,0.1}
\definecolor{darkgreen}{rgb}{0.1,0.7,0.1}
\definecolor{cyan}{rgb}{0.7,0.0,0.7}
\definecolor{dblue}{rgb}{0.2,0.2,0.8}
\definecolor{maroon}{rgb}{0.76,.13,.28}
\definecolor{burntorange}{rgb}{0.81,.33,0}
\definecolor{tealblue}{rgb}{0.212,0.459, 0.533}
\definecolor{darkyellow}{rgb}{0.6 , 0.5, 0.0}
\definecolor{mybrown}{rgb}{0.87058824, 0.56078431, 0.01960784}
\definecolor{myorange}{rgb}{0.835, 0.368, 0}
\definecolor{mygt}{rgb}{0.0078125 , 0.57421875, 0.40625}
\definecolor{mysp}{rgb}{0.84765625, 0.515625  , 0.0234375}
\definecolor{mypink}{rgb}{0.93359375, 0.62109375, 0.83984375}
\definecolor{mypurple}{rgb}{0.5372549 , 0.20588235, 0.99372549}
\definecolor{myblue}{HTML}{185ABC} 	
\definecolor{mygreen}{HTML}{137333}
\definecolor{carmine}{rgb}{0.59, 0.0, 0.09}
\definecolor{cornflowerblue}{rgb}{0.39, 0.58, 0.93}
\definecolor{brightmaroon}{rgb}{0.76, 0.13, 0.28}

\definecolor{pp}{rgb}{0.43921569, 0.18823529, 0.62745098}
\definecolor{rr}{rgb}{0.5254902 , 0.00784314, 0.12941176}
\definecolor{bb}{rgb}{0.0, 0.12549019607,0.57647058823}
\definecolor{yy}{rgb}{0.49803922, 0.3372549 , 0.0}
\definecolor{gg}{rgb}{0.02352941, 0.3372549 , 0.17647059}
\definecolor{highlightRowColor}{rgb}{0.95, 0.95, 1}

\newcommand{\bn}{{\bf n}}
\newcommand{\bv}{{\bf v}}
\newcommand{\bI}{{\bf I}}
\newcommand{\bG}{{\bf G}}
\newcommand{\bK}{{\bf K}}
\newcommand{\bR}{{\bf R}}
\newcommand{\bH}{{\bf H}}
\newcommand{\bt}{{\bf t}}
\newcommand{\bc}{{\bf c}}
\newcommand{\bC}{{\bf C}}

\newcommand{\bd}{{\bf d}}

\newcommand{\cP}{{\mathcal{P}}}
\newcommand{\cB}{{\mathcal{B}}}
\newcommand{\cF}{{\mathcal{F}}}

\ifdefined\ShowNotes
  \newcommand{\colornote}[3]{{\color{#1}\bf{#2: #3}\normalfont}}
\else
  \newcommand{\colornote}[3]{}
\fi

\newcommand{\eat}[1]{} 

\usepackage{tabularx}
\newcolumntype{C}{>{\centering\arraybackslash}X}
\newcolumntype{R}{>{\raggedleft\arraybackslash}X}
\newcolumntype{L}{>{\raggedright\arraybackslash}X}

\usepackage{amsmath,amsfonts,bm}

\def\1{\bm{1}}

\def\sp{space}

\DeclareMathAlphabet{\mathsfit}{\encodingdefault}{\sfdefault}{m}{sl}
\SetMathAlphabet{\mathsfit}{bold}{\encodingdefault}{\sfdefault}{bx}{n}

\makeatletter
\renewcommand{\paragraph}{%
  \@startsection{paragraph}{4}%
  {\z@}{0.4ex \@plus 1ex \@minus .1ex}{-1em}%
  {\normalfont\normalsize\bfseries}%
}
\makeatother

\usepackage[pagebackref,breaklinks,colorlinks, citecolor=ForestGreen]{hyperref}

\usepackage[capitalize]{cleveref}
\crefname{section}{Sec.}{Secs.}
\Crefname{section}{Section}{Sections}
\Crefname{table}{Table}{Tables}
\crefname{table}{Tab.}{Tabs.}

\def\cvprPaperID{5175} 
\def\confName{CVPR}
\def\confYear{2022}

\begin{document}

\title{Neural Volumetric Object Selection}
\author{
Zhongzheng Ren$^{1}\footnotemark{}$ \quad  Aseem Agarwala$^{2\dagger}$ \quad Bryan Russell$^{2\dagger}$ \quad Alexander G. Schwing$^{1\dagger}$ \quad Oliver Wang$^{2\dagger}$ \\
{$^{1}$University of Illinois at Urbana-Champaign \qquad $^{2}$Adobe Research } 
\\ {\normalsize \url{https://jason718.github.io/nvos}}
\vspace{-0.5cm}}

\twocolumn[{
\renewcommand\twocolumn[1][]{#1}
\maketitle
\begin{center}
\setlength{\itemheight}{2.4cm}
\setlength\tabcolsep{1.5pt}%
\begin{tabular}{ccccc}
\includegraphics[height=\itemheight]{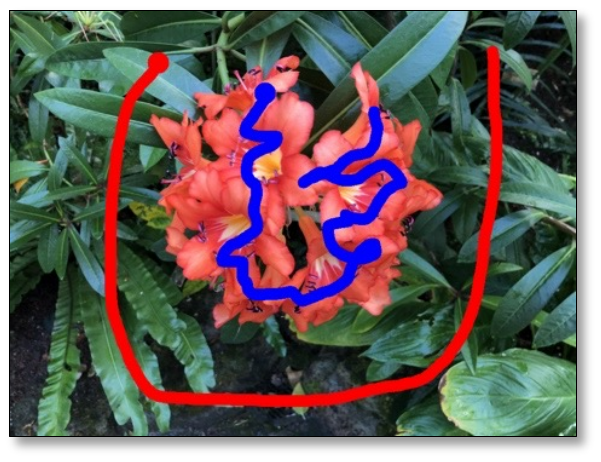} &
\includegraphics[height=\itemheight]{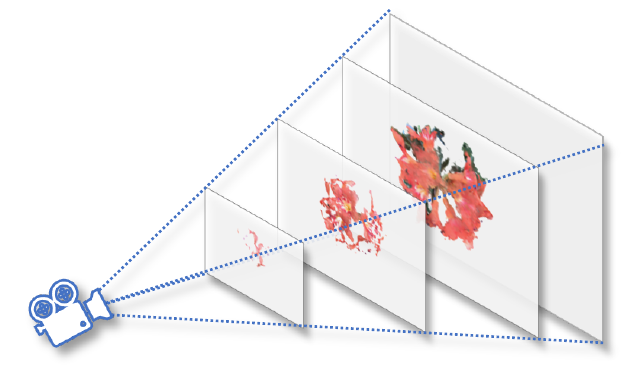} &
\multicolumn{1}{c:}{
\includegraphics[height=\itemheight]{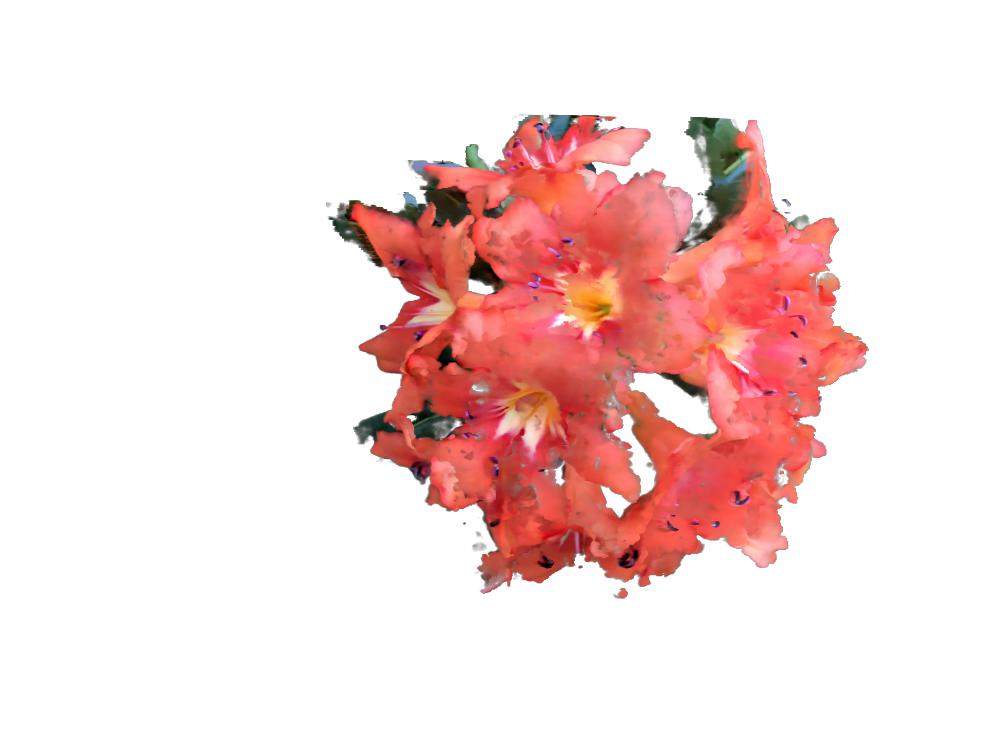}} &
\includegraphics[height=\itemheight]{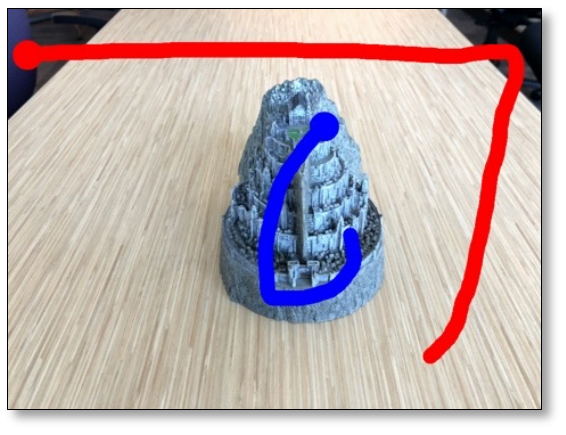}  &
\includegraphics[height=\itemheight]{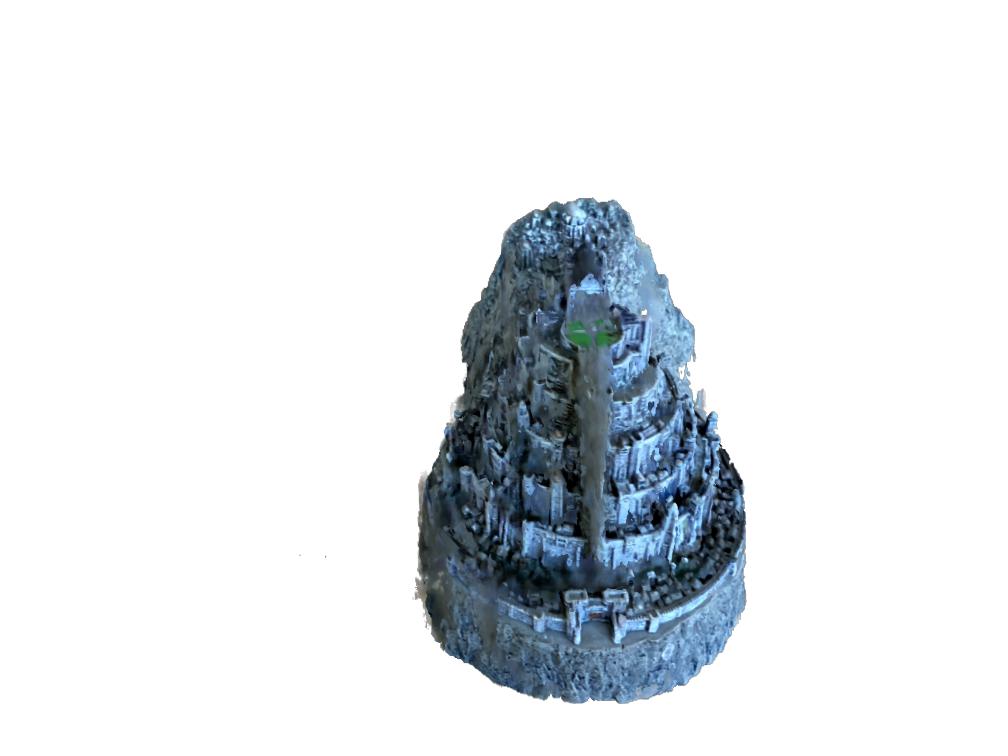} \\
\includegraphics[height=\itemheight]{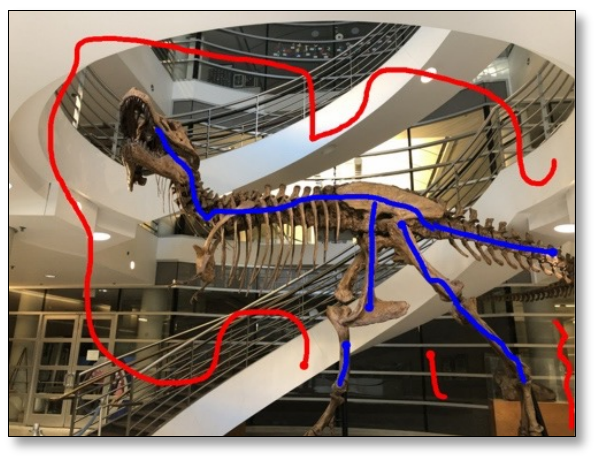}  &
\includegraphics[height=\itemheight]{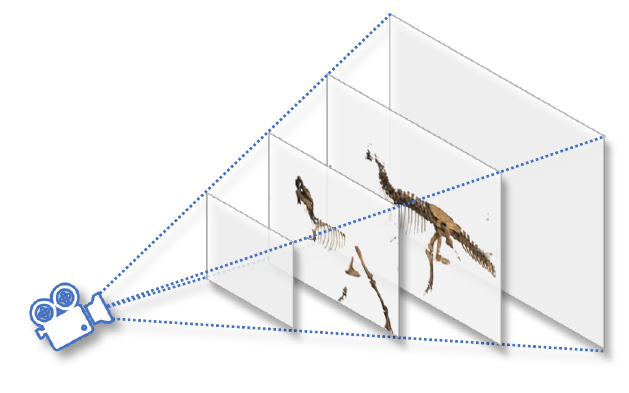} &
\multicolumn{1}{c:}{
\includegraphics[height=\itemheight]{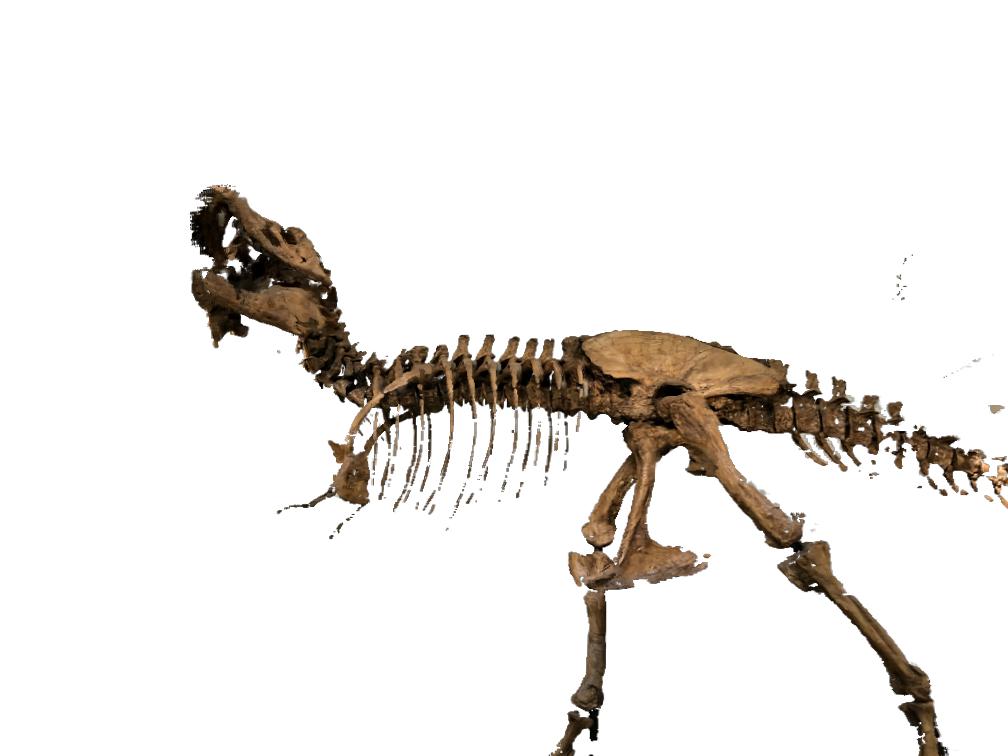}}  &
\includegraphics[clip,height=\itemheight]{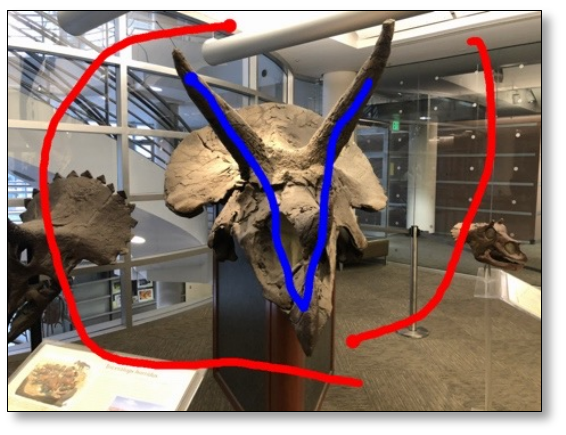}  &
\includegraphics[height=\itemheight]{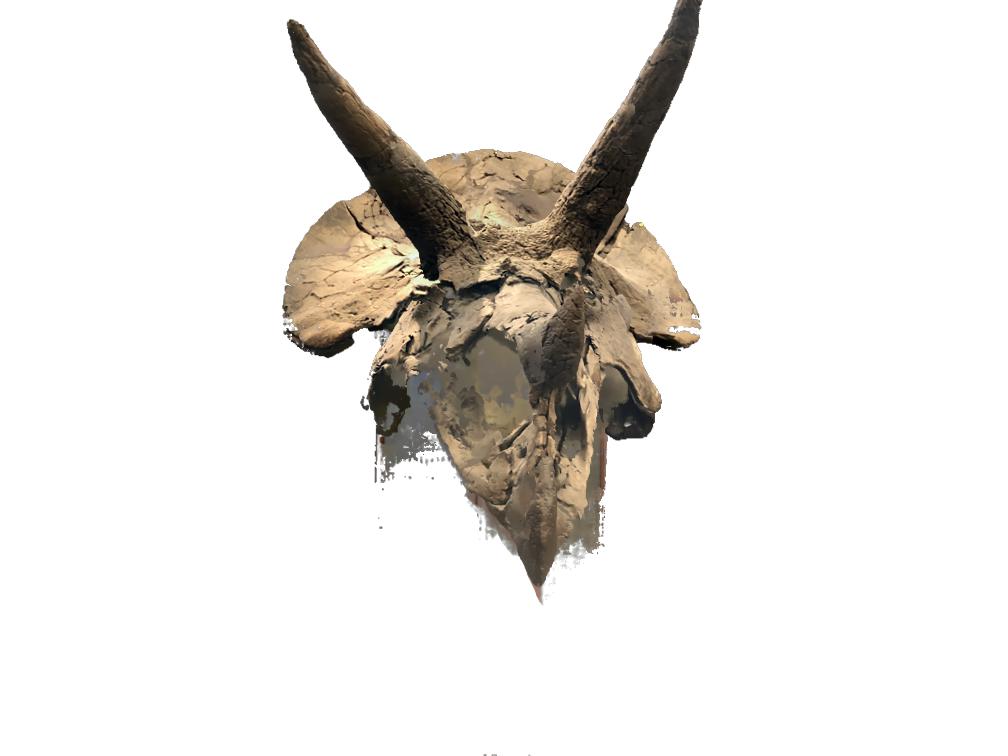}  \\
2D image \& scribble & 3D object selection & Novel view & 2D image \& scribble & Novel view 
\end{tabular}
\vspace{-0.3cm}
\captionof{figure}{We present an interactive 3D object selection method from neural volumetric representations (\eg, MPIs~\cite{ZhouTOG2018} or NeRFs~\cite{MildenhallECCV2020}).
{\color{bb} Blue}/{\color{rr} Red} scribbles denote {\color{bb} foreground}/{\color{rr} background}.
Please view with Adobe Acrobat or KDE Okular to see animations.
}
\label{fig:teaser}
\end{center}
}]

\begin{abstract}
We introduce an approach for selecting objects in neural volumetric 3D representations, such as multi-plane images (MPI) and neural radiance fields (NeRF). Our approach takes a set of foreground and background 2D user scribbles in one view and automatically estimates a 3D segmentation of the desired object, which can be rendered into novel views. To achieve this result, we propose a novel voxel feature embedding that incorporates  the neural volumetric 3D representation and multi-view image features from all input views. To evaluate our approach, we introduce a new dataset of human-provided segmentation masks for depicted objects in real-world multi-view scene captures. We show that our approach out-performs strong baselines, including 2D segmentation and 3D segmentation approaches adapted to our task. 
\end{abstract}

\makeatletter{\renewcommand*{\@makefnmark}{}
\footnotetext{$^*$Work partly done during an internship at Adobe Research.}
\footnotetext{$^\dagger$Alphabetic order.}
\makeatother}

\section{Introduction}
\label{sec:intro}
Object selection is an important task in an artist's workflow. With the growth of 3D photography, there is a need to support a suite of editing operations, such as object selection, that are readily available for 2D photographs. In this work, we consider object selection in neural volumetric 3D representations. Neural volumetric 3D representations---explicitly via Multi-plane Images (MPI)~\cite{szeliski1998stereo} or implicitly via Neural Radiance Fields (NeRF)~\cite{MildenhallECCV2020}---recover a remarkably accurate 3D representation of a scene from a given set of multi-view images. These representations have been shown to be particularly useful for novel-view synthesis~\cite{ZhouTOG2018, Wizadwongsa2021NeX, MildenhallECCV2020,ChenICCV2021,li2021neural,yu2021plenoctrees}, as this is the task that they are trained for. However, beyond just visualizing the same scene from a novel viewpoint, often we want to create a 3D reconstruction of an object so that we can extract it, and place it in a different context. There is, of course, substantial research in automatic object segmentation. However, user-driven methods, \eg, Photoshop, remain the most common means, as \emph{which} object to select is fundamentally a high-level decision. Similar user-driven methods for 3D object segmentation are particularly important for augmented and virtual reality (AR/VR) applications, \eg, if one wants to composite a selected object into a new scene, apply a filter to a selected object, remove a selected object, or share a real-world object with friends in an AR/VR environment.

While user-driven segmentation for 2D images has been studied for decades~\cite{FishkinCompGraph1984,AdamGeoPhys1986,MitsunagaSIGGRAPH1995,RuzonCVPR2000}, very little is known about how segmentation would work in those novel 3D scene representations. Specifically, for image segmentation, early works focus on energy minimization with graph cuts~\cite{BoykovICCV2001}, different forms of user interactions~\cite{Mortensen1995,KassIJCV1988,rother2004grabcut,BoykovICCV2001}, and ways to obtain better object priors~\cite{VekslerECCV2008,GulshanCVPR2010}. More recently, research has focused on encoding object priors with deep nets~\cite{XuCVPR2016,LiewICCV2017,LiCVPR2018,LiewICCV2019,LinCVPR2020}.

Naively applying these techniques to the set of images that are used to capture neural volumetric 3D representations is sub-optimal.
For example, simply transferring user interactions like scribbles from one image to another using a known camera transformation may fail to cover the intended object because of occlusions. Similarly, transferring an appearance model learned on one image to all remaining images is challenging because of appearance and lighting changes. Asking a user to interact with all images requires an interface that may not be intuitive or require excessive work, and furthermore may produce a segmentation that is not view-consistent.

For those reasons, novel user-driven 3D segmentation techniques are warranted. We propose a novel voxel feature embedding that incorporates  discretized features from the neural volumetric 3D representation and image features from all input views. Formally, we first project user interactions in the form of 2D scribbles from a reference image to sparse 3D locations using the reconstructed scene. We then learn a 3D object representation model that incorporates image features from all views via a developed multi-view feature embedding. We use these features to directly segment the object in the volumetric 3D scene representation and apply a post-processing step to remove outliers. The extracted 3D object can subsequently be viewed from different directions, as visualized in \figref{fig:teaser}.

We evaluate the proposed method on real world samples from the LLFF~\cite{mildenhall2019llff}, Shiny~\cite{Wizadwongsa2021NeX}, and NeRF-real360~\cite{MildenhallECCV2020} data. As shown in \figref{fig:teaser}, despite very few scribbles on a single reference image, the proposed method recovers an accurate 3D model of the object of interest and retains fine details (\eg, the ribs of the dinosaur in row 1). To study quantitatively, we obtained annotations using a professional service. Our method out-performs 2D and 3D interactive segmentation baselines by a  margin on all benchmarks.

In summary, we present the first method for user-driven 3D object selection targeting recent neural volumetric reconstruction.
We show that  a pre-trained network to embed multi-view features produces a more robust selection method than applying existing interactive 3D segmentation methods. For evaluation, we contribute a set of high-quality ground-truth annotations on three real-world datasets.
\section{Related work}
\label{sec:related}

\paragraph{2D interactive segmentation}
approaches include binary object segmentation or alpha matting, which aim to estimate the proportion of two colors mixing to form a boundary. Early matting work goes back to the 1980s~\cite{FishkinCompGraph1984,AdamGeoPhys1986} and  the 1990s~\cite{MitsunagaSIGGRAPH1995,RuzonCVPR2000}. Subsequent improvements like GrabCut~\cite{rother2004grabcut} used a global energy minimization and popularized  a simplification of the challenging matting task: first, estimate a ``hard'' segmentation; second, use border matting to compute alpha around a small strip at the boundary. A global energy minimization for interactive object segmentation has been used before by Boykov and Jolly~\cite{BoykovICCV2001} for object segmentation. Different forms of user input have been studied for interactive segmentation, including contours~\cite{Mortensen1995,KassIJCV1988}, bounding boxes~\cite{rother2004grabcut} and strokes~\cite{BoykovICCV2001,LiTOG2004,GradyPAMI2006,VekslerECCV2008,ren-eccv2020,GulshanCVPR2010,VezhnevetsGraphicon2005,CriminisiECCV2008,BaiIJCV2009,PriceCVPR2010}. 

Beyond different user interactions, follow-up work has also focused on improving the energy function employed in the work by Boykov and Jolly~\cite{BoykovICCV2001} and GrabCut~\cite{rother2004grabcut}. For example, Grady~\cite{GradyPAMI2006} studied random walks for speed-up and Veksler~\cite{VekslerECCV2008} and Gulshan \etal~\cite{GulshanCVPR2010} introduced shape priors to incorporate more expressive object information.

More recently, more expressive object information has been incorporated via deep nets~\cite{XuCVPR2016,LiewICCV2017,LiCVPR2018,LiewICCV2019,LinCVPR2020}. For instance, given user input, Xu \etal~\cite{XuCVPR2016} finetune fully connected nets (FCNs), the output of which is then used in a graph-cut formulation. Subsequent work refined the predictions by incorporating diversity or attention~\cite{LiewICCV2017,LiCVPR2018,LiewICCV2019,LinCVPR2020}. 

Different from the aforementioned works, we operate in a 3D volume rather than in image space. This setup requires developing a multi-view feature embedding which transforms scribbles from image space to volume space, and a pipeline that involves a 3D segmentation network, as well as a 3D graph-cut based post-processing step.

\begin{figure*}[t]
\centering
\includegraphics[width=\linewidth]{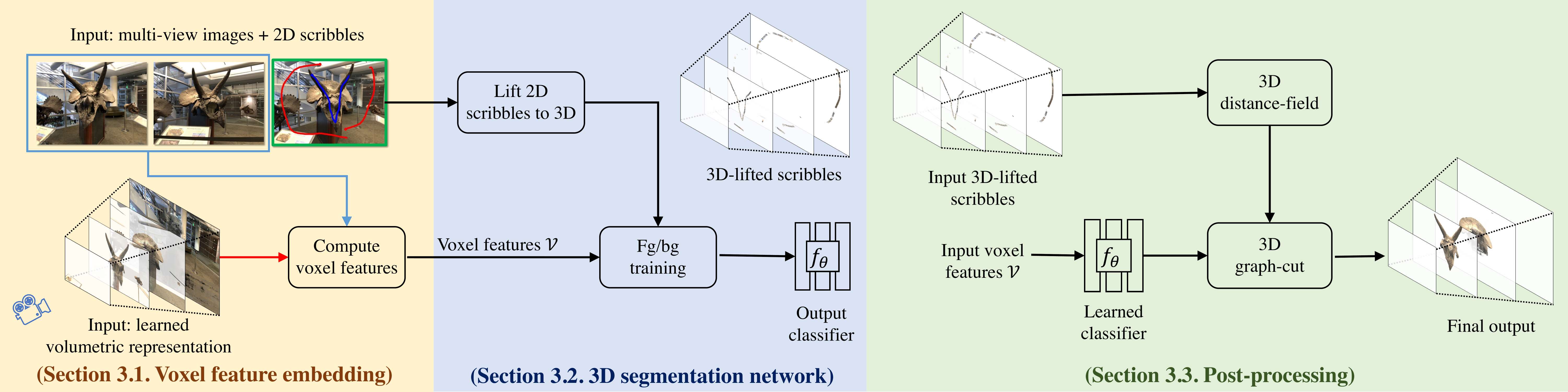}
\vspace{-0.7cm}
\caption{\textbf{Approach overview.} For each voxel in the 3D volume, we first compute a voxel feature embedding (\textbf{\color{darkyellow}left}). We then train a 3D segmentation network to classify each voxel into foreground or background using as supervision the partial user scribbles (\textbf{\color{bb}center}). We apply the learned classifier and further refine the result using a 3D graph-cut which uses the 3D distance field of the scribbles (\textbf{\color{gg}right}).}
\label{fig:archi}
\vspace{-0.5cm}
\end{figure*}

\paragraph{3D interactive segmentation} is particularly common in the medical community. Indeed early image segmentation techniques were often developed with medical image segmentation in mind~\cite{BoykovICCV2001,GradyPAMI2006}. Extending energy minimization techniques to 3D volumes proved to be challenging, necessitating various forms of improvements~\cite{GradyPAMI2006} to cope with increased memory requirements. Deep learning based techniques have been adopted more recently~\cite{WangPAMI2018,LiaoCVPR2020}, having the goal to capture object priors more accurately. Our focus is on visual realism instead of medical structure segmentation.

\paragraph{Image-based rendering (IBR)} has a long history in computer graphics and computer vision~\cite{ChenSIGGRAPH1993}, just like interactive segmentation. Image-based rendering aims to render novel views directly from input images and can be broadly categorized into methods which use geometry explicitly, methods which use geometry implicitly, and methods which do not use geometry at all. Classical techniques based on explicit geometry are texture-mapped models. Layered depth images (LDIs)~\cite{ShadeSIGGRAPH1998}, lumigraphs~\cite{BuehlerSIGGRAPH2001}, flow-based~\cite{ChenSIGGRAPH1993}, and tensor-based methods~\cite{AvidanCVPR1997} implicitly use geometry, while light-field methods~\cite{LevoySIGGRAPH1996} try to avoid using geometry. Hybrid methods~\cite{DebevecSIGGRAPH1996} have also long been studied.

More recently deep nets have been used for image-based rendering. Among the most widely used techniques are neural radiance fields (NeRF)~\cite{MildenhallECCV2020} and multi-plane images (MPI)~\cite{ZhouTOG2018}. Both have in common that they aim to extract from a given set of images a volumetric representation of the observed scene. While the volumetric representation is discrete for MPI and permits fast rendering, NeRF uses a continuous field to represent the volume.

Studying these representations for editing applications is an exciting direction. EditNeRF looked at modifying the colors and shape of a conditional radiance field given user scribbles~\cite{liu2021editing}. Guo \etal~\cite{guo2020objectcentric} looked at composing photo-realistic scenes of captured objects. Yang \etal~\cite{yang2021compositional} leverage 2D object instance masks to train a compositional model for repositioning objects in a scene. Our goal is to use this volumetric representation for interactive object segmentation given user scribbles. We think these representations are particularly useful for this application and permit more accurate recovery of the object than the classical 2D and 3D interactive segmentation techniques discussed before. We note a contemporary effort that obtains (non-interactive) object segmentation in synthetic scenes by using unsupervised learning for object radiance fields~\cite{yu2021unsupervised}.  Our work is the first that studies the use of these representations for interactive object selection in real-world scenes. 

\section{Approach}
\label{sec:approach}
Given a set of multi-view images of a scene, the corresponding 3D volumetric representation, and a pair of 2D scribbles indicating the foreground and background in one specific view (which we refer to as the reference view), our goal is to segment the foreground object in the 3D volumetric representation. In this paper, we use  3D volumetric representations discretized from neural IBR models, \ie, MPI and NeRF, due to their high quality view-synthesis results.

This problem is challenging as user-provided 2D scribbles need to first be translated to sparse and possibly inaccurate 3D annotations. Equally important, the neural 3D volume representation of IBR models is often noisy as its geometry is never explicitly supervised during training. Instead, the volume is learned in an implicit manner through differentiable volume rendering. We find a classical 3D graph-cut solution struggles with these neural 3D volumetric representations, often failing or requiring frequent scene-specific parameter tuning. To address this failure, we propose a robust and expressive voxel feature embedding that incorporates multi-view features from the neural volumetric 3D representation and from image features from all input views. We find this feature representation allows us to train a simple yet accurate classifier, which can then be refined with a small amount of post-processing via graph-cuts.

\paragraph{Overview.}
Our method operates on a discretized 3D volume $\cV = \{\bv_p\}$ where $\bv_p\in\mathbb{R}^{C}$ is the $C$-dimensional voxel feature embedding representing color and transparency of each 3D voxel location $p\in\mathbb{R}^3$. Given a set of 2D foreground and background scribbles in one specific view, we train a 3D segmentation network 
\begin{equation}
f_\theta(\bv_p): \mathbb{R}^C\rightarrow [0,1], 
\label{eq:network}
\end{equation}
which predicts the foreground probability of each voxel $p$ and which has learnable parameters $\theta$. We first illustrate the voxel features, which are composed of a new feature derived from the multi-view input training images, its appearance as predicted by the IBR model (\eg, color, density, and maybe view-directional components), and the location of the voxel~(\S~\ref{sec:feat}). We then introduce the 3D segmentation network $f_\theta$~(\S~\ref{sec:network}). Lastly, we introduce the post-processing refinement~(\S~\ref{sec:post}). We illustrate the overall method in Fig.~\ref{fig:archi}.

\subsection{Voxel feature embedding}
\label{sec:feat}
To conduct 3D segmentation with only sparse and noisy 3D supervision, we need an expressive voxel representation that captures both 3D location and appearance information. We develop a novel representation $\bv_p$ for each voxel $p$ in the volume, which is the  concatenation of three  features, \begin{equation}
\bv_p=[\bv_p^\text{MVS};\bv_p^\text{IBR}; \bv_p^\text{XYZ}],
\label{eq:feat}
\end{equation}
where $\bv_p^\text{MVS}$ is our novel multi-view image feature embedding, $\bv_p^\text{IBR}$ are discretized features extracted from IBR models, and $\bv_p^\text{XYZ}$ is a 3D positional encoding. We illustrate the process to obtain the features in the {\color{darkyellow}yellow} colored region of Fig.~\ref{fig:archi} and discuss it next. 

\paragraph{Multi-view image features $\bv_p^\text{MVS}$.}
The multi-view image features $\bv_p^\text{MVS}$ encode appearance information from the observed views. We find this information to be particularly useful for user-driven segmentation. This is intuitive when considering the three limitations of features that are extracted from only an IBR volume:  1) features extracted from an IBR volume are particularly noisy since the geometry is learned implicitly through differentiable volume rendering and never explicitly supervised. They further degrade during discretization; 2) IBR volume voxels encode limited neighborhood information; and 3) IBR volume representations are learned to model appearance, which might be sub-optimal for recognition tasks. To address these three limitations, we develop the multi-view image feature $\bv_p^\text{MVS}$ inspired by recent deep multi-view stereo (MVS) methods~\cite{yao2018mvsnet, ChenICCV2021}. Specifically, we encode multi-view images  into the reference view using the following three steps:

\noindent{$\bullet$ 2D feature extraction:}
for a multi-view image set $\{\bI_i\}_{i=1}^M$ where $M$ denotes the  number of available views, a pre-trained 2D convolutional neural network (CNN) is used to extract image features $\{\bG_i\}_{i=1}^M$ for each image $i$.

\noindent{$\bullet$ Cost-volume construction:} 
using the  known camera intrinsic and extrinsic parameters $\{\bK_i, \bR_i, \bt_i\}_{i=1}^M$, the extracted 2D feature maps $\{\bG_i\}_{i=1}^M$ can be warped onto multiple 3D planes oriented fronto-parallel to the reference view $r\in\{1,\dots,M\}$, and their agreement can be recorded in a plane-sweep cost volume. Let $(u,v)$ be a pixel location in the reference view $r$ and $\bH_{i\rightarrow r}(z)$ be a $3\times 3$ homography matrix that projects a 2D homogeneous point in view $i$ to the plane at depth $z$ of reference view $r$. We obtain the warped feature map $\bG_{i\rightarrow r, z}$ by applying the homography matrix to the pixel locations of the feature map $\bG_i$ for view $i$,  
\begin{equation}
\bG_{i\rightarrow r, z}(u, v)= \bG_i\left(\bH_{i\rightarrow r}^{-1}(z)[u, v, 1]^T\right).
\label{eq:feat-warp}
\end{equation}
If $\bn_r$ is the principal axis of the reference camera $r$, then the homography $\bH_{i\rightarrow r}(z)$ is:
\begin{equation}
\bH_{i\rightarrow r}(z) = \bK_i \bR_i \left(\bI - \frac{(\bt_r-\bt_i)\bn_r^T}{z}\right) \bR_r^T \bK_r^T.
\label{eq:homo}
\end{equation}

We then aggregate the projected features from the $M$ plane-sweep volumes into a single cost-volume via 
\begin{equation}
\bG_{r, z}(u, v) =  \text{Var}_{i\in\{1, \dots, M\}}(\bG_{i\rightarrow r, z}(u, v)), 
\label{eq:cost-vol}
\end{equation}
where $\text{Var}(\cdot)$ calculates the variance of the features over the $M$ feature maps.
We calculate the variance as it explicitly measures the feature difference from multiple views, which has been validated to out-perform baselines calculating the features' mean~\cite{yao2018mvsnet}.
The computed visual feature variance serves as a good indicator for probable surface locations and hence greatly informs the 3D segmentation.

\noindent{$\bullet$ Feature computation}: the computed cost-volume is often noisy due to occlusions or non-Lambertian reflectance. 
Therefore, we further refine it using a 3D U-Net~\cite{ronneberger2015u}. Concretely, we compute the final feature volume via
\begin{equation}
\bv_{(\cdot)}^\text{MVS} = g_\omega(\bG_{r}),
\label{eq:3d-unet}
\end{equation}
where we concatenate each $\bG_{r, z}$ along the Z-axis to form the 3D tensor $\bG_r$ and $g_\omega$ is a 3D U-Net with parameters $\omega$.  To extract robust and expressive multi-view  features, we adopt the learned weights $\omega$ from Chen~\etal~\cite{ChenICCV2021}, who originally train the network $g_\omega$ on all scenes of the DTU dataset~\cite{jensen2014large} for fast generalization of NeRF models to unseen scenes. Appendix Fig.~\ref{fig:mvs-net} shows the details for the computation of $\bv_p^\text{MVS}$.

\paragraph{Neural voxel features $\bv_p^\text{IBR}$.}
We also extract voxel features from the neural IBR volume.
To study the robustness and generalizability of  our method, we use two recent IBR models: MPI and NeRF. An MPI is naturally a discretized volume for 3D scenes and we use the MPI variant NeX~\cite{Wizadwongsa2021NeX} here. In contrast, a NeRF is an implicit continuous neural representation and cannot be directly used. We hence adopt the PlenOctrees~\cite{yu2021plenoctrees} discretization which is an octree-based representation that supports real-time rendering without compromising photometric quality. PlenOctrees convert a NeRF model to a regular volume of size $512^3$. For the obtained volumes, both NeX and PlenOctrees leverage spherical basis functions for modeling the color information via
\begin{equation}
    c_p(\bd) = \sum_{l\in\{1,\dots,N\}} k_p^l H^l(\bd), 
\end{equation}
where $c_p(\bd)\in\mathbb{R}^3$ is the color of voxel $p$ from view direction $\bd$, $k_p^l\in\mathbb{R}^3$ for voxel $p$ are RGB coefficients, $H^l(\bd)$ is a view-dependent basis function, and $l\in \{1, \dots, N\}$ is the basis function index. In addition, each voxel also stores one transparency value which we refer to as $\xi_p$ (\eg, alpha-transparency  in an MPI and density in a NeRF). The neural IBR feature is constructed as,
\begin{equation}
    \bv_p^\text{IBR} = [\xi_p, k_p^1, \dots, k_p^N], 
\end{equation}
where $[\cdot]$ denotes concatenation. Note that this feature is independent of the viewing direction $\bd$.

\paragraph{Positional voxel features $\bv_p^\text{XYZ}$.}
In addition to the multi-view image features, we also extract a positional encoding of the voxel location. For each voxel, we project its $(x,y)$ location to a 40-dimensional feature vector using a positional encoding~\cite{MildenhallECCV2020}, and similarly its $z$ location to a 16-dimensional feature vector. In total, for each voxel we obtain a 56 dimensional feature vector  $\bv_p^\text{XYZ}$.

\subsection{3D segmentation network}
\label{sec:network}
Given the  voxel representation $\bv_p\in\mathbb{R}^C$ detailed in \S~\ref{sec:feat}, we use a 3-layer MLP as the segmentation network $f_\theta$ to predict the foreground probability. 

As user input is provided in the form of 2D scribbles on the reference view, to obtain training labels, we first lift the 2D foreground and background scribbles to 3D using the known camera pose. For this lifting, we define a 3D ray for each pixel and compute the intersecting 3D surface-voxel as the first voxel on the ray with accumulated transmittance lower than  $\gamma=0.01$. 
This step yields an ``expected depth''  for each ray, which permits to assign to the corresponding voxel either a foreground or a background label. Our network is then trained for binary classification using a standard binary cross-entropy loss. This process is illustrated in the {\color{bb}blue} colored region of Fig.~\ref{fig:archi}.

\subsection{Post-processing}
\label{sec:post}

Since the 3D segmentation network (\S~\ref{sec:network}) is trained with limited supervision (sparse scribbles), the final prediction on the entire volume is occasionally noisy. There are two main causes for the noise: 1) \textbf{floaters}: these errors are generally isolated and far from the foreground scribbles in 3D space; and 2) \textbf{incompleteness}: surface voxels are predicted incorrectly if their appearance differs significantly from the foreground scribbles, even though these voxels are very close to the foreground scribbles in 3D space. The reason for these errors is that the classifier processes each voxel independently, \ie,  neighborhood correlations are not considered. To address this issue, we apply a distance field-based 3D graph-cut for post-processing. Note, the neural volumetric representations are often of high resolution, which graph-cut fails to process due to memory and computational limitations. To ensure fast inference, we operate in a down-sampled and truncated volume space and upsample the segmentation afterwards. Truncation removes planes that are unlikely to contain the object of interest. We down-sample the volume by 4$\times$ on the XY-plane and truncate to 20 planes. Note that our initial 3D segmentation is in the original high-resolution volume.

We design the following energy function for 3D segmentation. As before, let $p$ be a voxel in the 3D volume and $N\subseteq |\cV|\times |\cV|$ be a neighborhood system on the volume (we adopt a 6-connected neighborhood). We seek to infer the foreground/background label $y_p\in \{0,1\}$ of each voxel $p$. Let the unary term $\phi_p$ indicate how likely a voxel $p$ belongs to foreground or background and the pairwise term $\psi_{p,q}$ capture the correlation between voxel $p$ and its neighboring voxel $q$. We minimize the sum of unary and pairwise terms,
\begin{equation}
 E = \sum_{p\in\cP} \phi_p(y_p) + \alpha \sum_{p,q\in N}\psi_{p,q}(y_p, y_q),
 \label{eq:energy}
\end{equation}
where $\alpha$ is a scalar balancing weight. We depict this process in the {\color{gg}green} colored region of Fig.~\ref{fig:archi}. 

The unary term is based on our network prediction and input scribbles, \ie, 
\begin{equation}
\phi_p(y_p) = \omega_1\phi_p^c(y_p)  + \omega_2 \phi_p^d(y_p). 
\label{eq:unary}
\end{equation}
Here, the first term relies on the segmentation network output and is formulated as
\begin{equation}
\phi_p^c (y_p) =
\left\{
	\begin{array}{ll}
		f_\theta(\bv_p) & \mbox{if } y_p = 1 \\
		1-f_\theta(\bv_p) & \mbox{if } y_p = 0
	\end{array}
\right.,
\label{eq:1st-unary}
\end{equation}
which is the probability that a voxel $p$ belongs to category $y_p$. The second term $\phi_p^d$ is based on a distance-field of voxel $p$ to the input scribbles. Formally,
\begin{equation}
\phi_p^d(y_p) =
\left\{
	\begin{array}{ll}
		\min_{q\in\cF}\;\text{Dist}(p,q) & \mbox{if } y_p = 1 \\
		\min_{q\in\cB}\;\text{Dist}(p,q) & \mbox{if } y_p = 0
	\end{array}
\right.,
\label{eq:dist-field}
\end{equation}
where $\cB, \cF$ are the set of 3D-projected background and foreground scribbles and $\text{Dist}(\cdot)$ is a function computing the 3D distance between two voxels.

The pairwise term $\psi_{p,q}$ models correlation. For instance, if the feature encoding of two voxels are similar and if both voxels are close, we expect them to be labeled similarly. This correlation is formulated via the binary term  
\begin{equation}
\psi_{p,q}(y_p,y_q) = |y_p - y_q| \cdot \text{Dist}(p,q)^{-1}\cdot\exp^{-\frac{(\bv_p-\bv_q)^2}{\sigma}},
\label{eq:binary}
\end{equation}
where $\sigma$ is a  positive scalar. This term is useful as the depth planes in the 3D volume may be irregularly spaced, \eg, inverse depth spacing is also used for real-world forward facing scenes~\cite{Wizadwongsa2021NeX}.

\section{Experiments}
\label{sec:exp}
In this section, we introduce our experimental setup and show quantitative results, followed by ablation studies and an analysis. Lastly, we provide a qualitative comparison.

\paragraph{Experimental setup.}
There are three stages: 
\textbf{1) training:} The classifier $f_\theta$ operates on only the features of voxels belonging to 3D-lifted scribbles, and is trained for binary classification (fg/bg). We annotate only one pair of scribbles (fg/bg) in the reference view.
\textbf{2) inference:} The trained $f_\theta$ classifies all voxels in the entire 3D volume. No scribbles are needed.
\textbf{3) evaluation:} with all voxels classified, we render the foreground voxel locations to a novel view for 2D evaluation (segmentation, photo-realism) relative to GT.

\paragraph{Datasets.}
We use three classical multi-view scene datasets with multiple objects. We test the MPI models on seven scenes from LLFF~\cite{mildenhall2019llff}, which are front-facing real-world scenes, with 20-62 images each. We also test on four scenes from  Shiny~\cite{Wizadwongsa2021NeX}, which is captured in a similar manner as LLFF but contains more challenging view-dependent effects such as metallic and transparent objects. We test NeRF-based models (PlenOctrees~\cite{yu2021plenoctrees}) on two real-world 360$^{\circ}$ scenes from NeRF~\cite{MildenhallECCV2020} (NeRF-real360). Unfortunately, PlenOctrees does not generalize well to the front-facing datasets LLFF and Shiny due to the large scene depth range. We thus leave the task of 3D segmentation using NeRF models on front-facing scenes to future work. For all  datasets, the input images are resized to a $1008\times 756$ pixels following the common practice in novel-view synthesis~\cite{MildenhallECCV2020, Wizadwongsa2021NeX}. The camera parameters are estimated via Structure-from-Motion (SfM) using the publicly available COLMAP library~\cite{schoenberger2016sfm}.

\begin{table}[t]
\setlength{\tabcolsep}{5pt}
\centering
\small{
\begin{tabular}{l|cc|cc}
\specialrule{.15em}{.05em}{.05em}
Dataset & \multicolumn{2}{c|}{LLFF} & \multicolumn{2}{c}{Shiny}\\
Metrics & Acc.$\uparrow$ & IoU$\uparrow$  & Acc.$\uparrow$ & IoU$\uparrow$\\
\hline
Random & 50.0 & 13.5 &50.0 & 13.5 \\
Scribbles (projected) & 8.1 & 16.6 & 8.0 & 20.7 \\
\hline
\rowcolor{highlightRowColor} \multicolumn{5}{c}{\textit{2D segmentation (using projected scribbles)}} \\
\hline
Graph-cut~\cite{kolmogorov2004energy} & 88.6 & 59.0 & 86.1 & 48.0 \\
GrabCut~\cite{rother2004grabcut} & 78.1 & 49.0 & 66.2 & 31.1 \\
DeepLabV3~\cite{chen2017rethinking} & 91.5 & 56.6 & 88.6 & 50.4 \\
DEXTR~\cite{Man2018dextr} & 89.7 & 34.5 &  59.6 & 40.4\\
FCA-Net~\cite{lin2020interactive} & 88.3 & 62.7 & 87.9 & 58.5 \\
\hline
\rowcolor{highlightRowColor} \multicolumn{5}{c}{\textit{3D segmentation}} \\
\hline
Graph-cut (3D) &73.6 & 39.4& 78.3&32.4 \\
Ours & \textbf{92.0} & \textbf{70.1} & \textbf{90.7} & \textbf{69.3} \\
\specialrule{.15em}{.05em}{.05em}
\end{tabular}}
\vspace{-0.3cm}
\caption{{\bf 2D mask evaluation results.} Our method yields more accurate object selection results on both datasets.}
\label{tab:res-mask}
\vspace{-0.4cm}
\end{table}

\paragraph{Scribbles.}

For evaluation, we annotate a fixed set of foreground-background scribbles per image as shown in Appendix Fig.~\ref{fig:scribble-llff} left column. 
For each scene, we have one pair of foreground ($\cF$) and background ($\cB$) scribbles for the reference view. Note that a scribble ($\cF$ or $\cB$) may contain several strokes.
We use the scribbles as the input to our method. For baselines, we project these scribbles to the testing view using the recovered camera (see details in \S~\ref{sec:network}). To test the generalizability, we ensure that the scribbles have different length and shape, sometimes covering multiple objects, \eg, `tools' and `pasta' in the Shiny dataset.

\paragraph{Evaluation annotations \& criteria.}
Our goal is to accurately segment 3D objects in volumetric representations. However, there is no standard way to annotate ground-truth 3D masks for real-world scenes since different IBR models represent 3D scenes differently. We propose two ways.

First, we  evaluate the projected 2D mask segmentation results. For  evaluation, we obtain high-quality 2D annotation masks in unseen validation views for the aforementioned three datasets using a professional image segmentation service. For each scene, one unseen validation image is annotated for evaluation. Note that the validation image  view differs from the  MPI reference view. 
Next, we render 2D foreground masks from the 3D volumetric representation in the novel view and compare  to the corresponding 2D ground-truth. We report pixel classification accuracy (Acc) and foreground intersection-over-union (IoU).

Second, as our method operates on volumetric 3D representations, the segmented foreground object could be rendered to novel views through volume rendering. We thus report the photo-realistic quality of the segmented object rendered in a novel view with a black background. Since the objects could be small and we do not want the background to dominate the evaluation. To overcome this issue, we crop  the foreground region on the rendered and ground-truth images using a tight bounding box around the ground-truth mask. We report Structural Similarity Index Measure (SSIM), Peak Signal-to-Noise Ratio (PSNR), and Learned Perceptual Image Patch Similarity (LPIPS)~\cite{zhang2018perceptual}. We do not report these metrics for the 2D baselines as they do not render new views.

\begin{figure*}[t]
\centering
\includegraphics[width=0.98\linewidth]{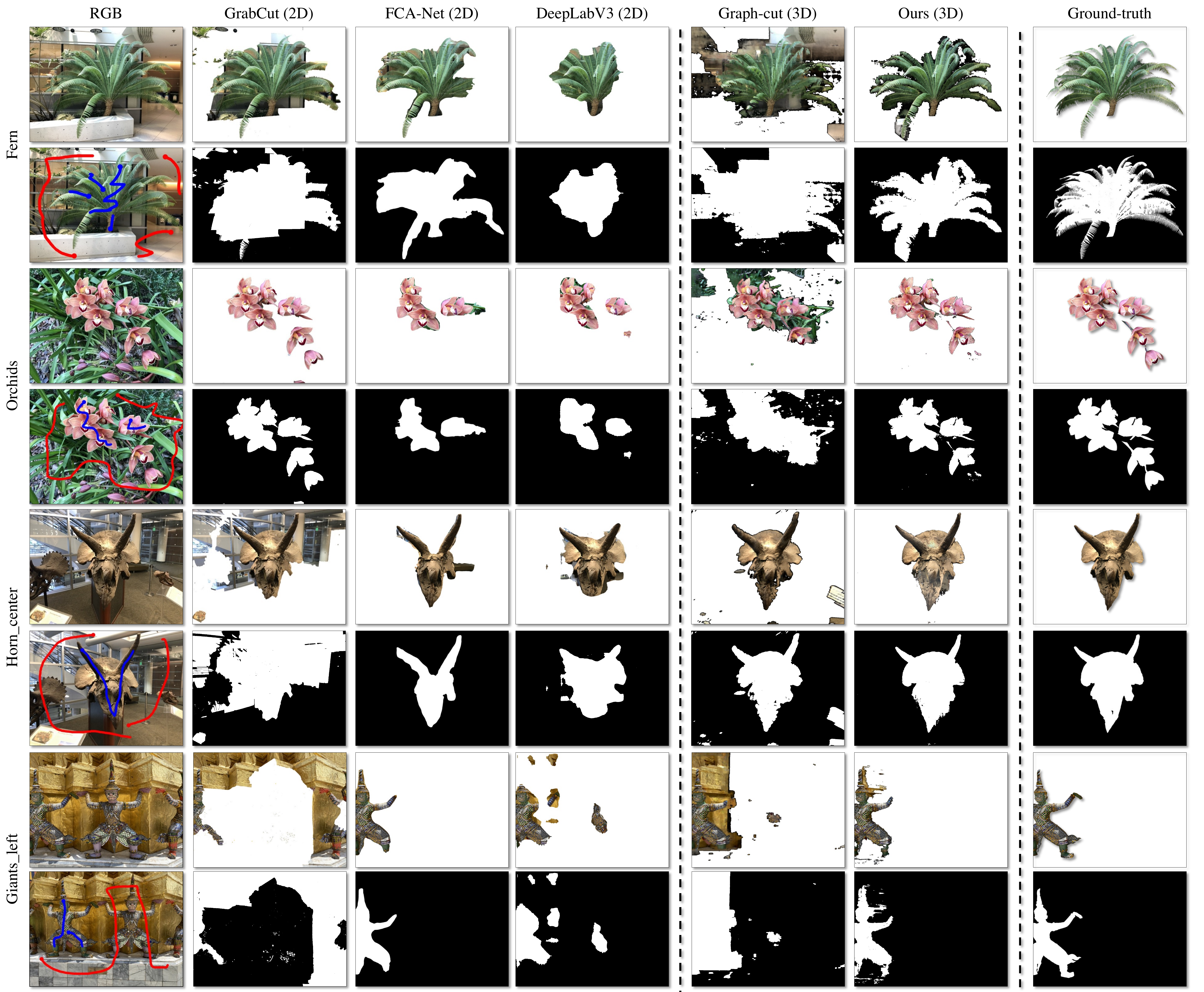}
\vspace{-0.3cm}
\caption{\textbf{Qualitative comparison.} For each scene, we show the reference view image and the input scribbles on the left. We then show the 2D mask and corresponding foreground image computed by different methods. On the right-most column, we show the ground-truth. Note that the foreground segmentation of the 2D baselines are not rendered; we apply the inferred mask to the ground truth test image.}
\label{fig:main}
\vspace{-0.4cm}
\end{figure*}

\begin{table}[t]
\setlength{\tabcolsep}{5pt}
\resizebox{\linewidth}{!}{
\centering
\begin{tabular}{l|ccc|ccc}
\specialrule{.15em}{.05em}{.05em}
Dataset & \multicolumn{3}{c|}{LLFF} & \multicolumn{3}{c}{Shiny}\\
Metrics  & SSIM$\uparrow$ & PSNR$\uparrow$ & LPIPS$\downarrow$ & SSIM$\uparrow$  & PSNR$\uparrow$  & LPIPS$\downarrow$ \\
\hline
Graph-cut (3D) & 0.600 & 15.03 &  0.415 & 0.454 & 12.83 & 0.477\\
Ours & \textbf{0.767} & \textbf{18.40} & \textbf{0.213} & \textbf{0.612} & \textbf{15.73} & \textbf{0.319}\\
\specialrule{.15em}{.05em}{.05em}
\end{tabular}}
\vspace{-0.3cm}
\caption{{\bf Novel-view object rendering results.} Our model renders more realistic foreground objects than 3D graph-cut.}
\label{tab:res-render}
\vspace{-0.4cm}
\end{table}

\paragraph{Baselines.}
We compare  to several baselines: 
1) \textbf{Random}: we randomly assign pixels to the two classes with equal probability.
2) \textbf{Scribbles (proj.)}: we  lift the input 2D scribbles from the reference view into 3D and find the intersecting voxels (see details in \S~\ref{sec:network}). We then project these voxels into the test image using the  camera matrices. Since the projected scribbles are used as the input of baseline methods, this experiment helps to understand how accurate the scribbles are after projection. We visualize these projected scribbles in Appendix Fig.~\ref{fig:scribble-llff} right column.

Since the evaluation is conducted in 2D, we consider 2D interactive segmentation  baselines. Specifically, given the input scribbles lifted to 3D and projected into the view that we have supervision for, we evaluate:
\textbf{1) Graph-cut~\cite{kolmogorov2004energy}}: we use the 2D Graph-cut with the unary term proposed in LazySnapping~\cite{li2004lazy} and with an exponential binary term. The exact formulation is provided in the Appendix.
\textbf{2) GrabCut~\cite{rother2004grabcut}}: an improved version of Graph-cut based on iterative energy minimization.
\textbf{3) DeepLabV3~\cite{chen2017rethinking}}: we fine-tune a state-of-the-art semantic segmentation net (DeepLabV3) for binary classification  using the projected scribbles.
\textbf{4) DEXTR~\cite{Man2018dextr}}: a pre-trained image interactive segmentation model that takes the 4 extreme points of the projected scribbles as input.
\textbf{5) FCA-Net~\cite{lin2020interactive}}: a pre-trained  2D interactive image segmentation model that takes user clicks as input. 

In contrast to 2D baselines, we aim to achieve 3D segmentation in volumetric representations rather than simply segmenting 2D objects in novel views. As a baseline for 3D interactive segmentation, we consider \textbf{Graph-cut (3D)}.
Concretely, we keep everything the same as in \S~\ref{sec:post} except changing the first unary term (\equref{eq:1st-unary}) to 
\begin{equation}
\phi_p^c (y_p) =
\left\{
	\begin{array}{ll}
	    \min_{q\in\cF}\;\|\bv_p^\text{IBR}-\bv_q^\text{IBR}\|_2 & \mbox{if } y_p = 1 \\
		\min_{q\in\cB}\;\|\bv_p^\text{IBR}-\bv_q^\text{IBR}\|_2 & \mbox{if } y_p = 0
	\end{array}
\right.,
\label{eq:3d-gc}
\end{equation}
where $\bv_{\cdot}^\text{IBR}$ is given  in \equref{eq:feat}.

\paragraph{Implementation details.}
The input images for both MPI and NeRF are resized to $1008\times756$ resolution. Following~\cite{Wizadwongsa2021NeX, yu2021plenoctrees}, the MPI volume is of dimension $1156\times 1408\times 192$ and the PlenOctree volume is of size $512^3$. Our cost volume is of size $640\times 960\times D$ where $D\in\{192, 512\}$ for MPI and PlenOctree, following MVS-Net~\cite{yao2018mvsnet}. Our classifier is implemented as a 2-layer MLP with hidden dimension 128. We train our network using the Adam optimizer with an initial learning rate of 0.001. During training, we adopt cross-validation where 10\% of the scribbles' voxels are used as a validation set. Once the hyper-parameters are selected, we re-train our network with all scribbles' voxels to obtain the final model. For the 3D graph-cut parameters, we use $w_1=1, w_2=10, \alpha=0.1, \sigma=1$. Please refer to Appendix~\S~\ref{appendix:impl} for more implementation details.

\begin{figure*}[t]
\centering
\includegraphics[width=\linewidth]{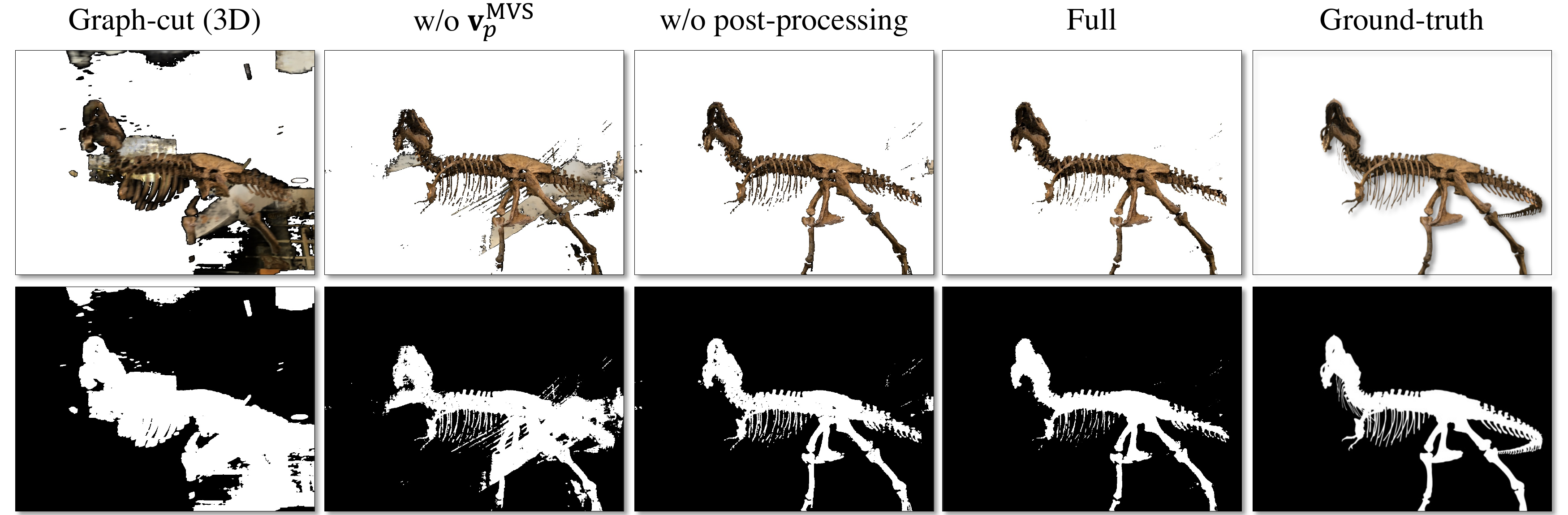}
\vspace{-0.5cm}
\caption{\textbf{Visual ablation.} Our method predicts more accurate and cleaner results than 3D segmentation baselines. Both the multi-view image feature $\bv_p^\text{MVS}$  and the post-processing modules contribute to the final performance.}
\vspace{-0.4cm}
\label{fig:ablation}
\end{figure*}

\subsection{Quantitative results}
\label{sec:quanti}
\noindent\textbf{2D mask evaluation.} 
We report the 2D mask evaluation results in Tab.~\ref{tab:res-mask}. We observe: 
1) our model achieves higher segmentation accuracy than the best-performing optimization-based 2D interactive segmentation methods (Graph-cut or Grabcut) by 11.1\%/21.3\% IoU on the LLFF/Shiny datasets;
2) our model also improves upon the best-performing deep learning-based 2D interactive segmentation methods (DEXTR or FCA-Net) by 7.4\%/10.8\% IoU;
3) the 2D supervised segmentation model (DeepLabV3) makes accurate prediction (91.5\%/88.6\% Acc), but has significantly lower IoU (56.6\%/50.4\%) than ours (70.1\%/69.3\%). 
4) our method also out-performs the 3D graph-cut baseline since it can only operate on a down-sampled volume and is not consistent across scene-specific neural volumetric representations. 

\paragraph{Novel-view object rendering.}
Our model segments the foreground object in 3D and thus is able to render the object in novel views. We report the rendering results in Tab.~\ref{tab:res-render} where we observe that our method significantly improves upon the 3D graph-cut baseline (+0.167 SSIM/+3.37 PSNR/-0.202 LPIPS on LLFF, and +0.158 SSIM/+2.90 PSNR/-0.158 LPIPS on Shiny).

\subsection{Ablation study}
\label{sec:analysis}

\begin{wraptable}{r}{0.25\textwidth}
\vspace{-0.3cm}
\resizebox{\linewidth}{!}{
\begin{tabular}{c|cc}
\specialrule{.15em}{.05em}{.05em}
Metrics & Acc$\uparrow$ & IoU$\uparrow$ \\
\hline
w/o $\bv_p^\text{MVS}$ & 87.6 & 64.7 \\
w/o $\bv_p^\text{IBR}$ & 89.1 & 60.4 \\
w/o $\bv_p^\text{XYZ}$ & 85.9 & 53.2 \\
\hline
Kinetics 3D CNN & 83.2 & 53.9\\
w/o post-proc. & 90.9 &  68.0  \\
\hline
Ours & \textbf{92.0} & \textbf{70.1} \\
\specialrule{.15em}{.05em}{.05em}
\end{tabular}}
\vspace{-0.3cm}
\caption{{\bf Ablation study on LLFF.}}
\vspace{-0.3cm}
\label{tab:res-ablation}
\end{wraptable}
\paragraph{How important is each feature?}
One of our contributions is the voxel feature embedding  detailed in \S~\ref{sec:feat} which consists of three different terms (\equref{eq:feat}).
We validate the effectiveness of each one of them in Tab.~\ref{tab:res-ablation} (top section). We observe that: 
1) removing the introduced multi-view image feature embedding $\bv_p^\text{MVS}$ hurts the performance by 5.4\% IoU; 
2) removing the learned feature $\bv_p^\text{IBR}$ hurts the performance by 9.7\% IoU; 
3) removing the positional encoding $\bv_p^\text{XYZ}$ yields the biggest IoU drop (16.9\%).

\paragraph{How important is the 3D U-Net?}
In our experiments, we use the DTU~\cite{jensen2014large} pre-trained 3D U-Net from Chen~\etal~\cite{ChenICCV2021} in \equref{eq:3d-unet}. We verify the effectiveness of this pre-trained 3D U-Net by replacing it with a Kinetics~\cite{kay2017kinetics} pre-trained 3D CNN encoder~\cite{tran2018closer}. As shown in Tab.~\ref{tab:res-ablation}, replacing the U-Net hurts the results (-8.8\% Acc, -16.2\% IoU).  

\paragraph{How important is the post-processing?}
We validate the effectiveness of post-processing in Tab.~\ref{tab:res-ablation} where we find removing it degrades the results by 2.1\% IoU. 

\subsection{Qualitative results}
\label{sec:quali}
We present qualitative results in Fig.~\ref{fig:main}. Compared to both the 2D and 3D baselines, we observe our method to predict more accurate and complete foreground masks across  objects. In addition, we also observe the proposed method to render foreground objects at testing views with good quality. 

We further illustrate the effectiveness of the multi-view feature $\bv_p^\text{MVS}$ and the post-processing in Fig.~\ref{fig:ablation}. Our method recovers fine-grained local details (\eg, ribs) and significantly out-performs 3D graph-cut. The multi-view image feature $\bv_p^\text{MVS}$ helps to better isolate the foreground object and post-processing helps visual smoothness, \ie, it removes floaters.

\section{Discussion}
\paragraph{Limitations.}
We observe the proposed method to struggle in two main cases: 
1) the final predictions still contain floaters (\eg, Fig.~\ref{fig:main} `Orchids' and `Giants\_left').
2) our voxel feature may not capture view-dependent effects like reflections (\eg, Fig.~\ref{fig:main} `Horns\_center' foreground image). 
Moreover, presently, our model does not run at interactive rates. While the features can be computed off-line, the segmentation network operates on the high-resolution volume which takes about 1-3min using our un-optimized implementation. The post-processing takes another 3-5min using un-optimized code. 
Faster feedback would make it easier for a user to determine where to add strokes. Finally, while our selections are currently binary, producing soft alpha mattes is an interesting area for future work.

As with most creative tools, object selection for compositing applications could be used for nefarious purposes, \eg, to misinform. To combat this, we suspect that both forensics tools as well as a general increased cultural understanding of threats are needed to mitigate potential damages.

\paragraph{Conclusion.}
We present a user-driven way to segment objects in neural volumetric representations using user input on a single frame. For this, we develop robust features that can be classified accurately with a neural network. We further improve results with graph-cuts post-processing. We find the proposed method handles  a variety of scenes well while scaling to high-resolution volumetric datasets. We believe that interactive segmentation in IBR volumes will be a key workflow in 3D asset generation and editing.

\noindent\textbf{Acknowledgements:} This work is supported in part by NSF \#1718221, 2008387, 2045586, 2106825, MRI \#1725729, NIFA 2020-67021-32799.

\clearpage
{\small
\bibliographystyle{ieee_fullname}
\bibliography{egbib}
}

\clearpage
\newcommand{\beginsupplement}{
    \setcounter{table}{0}
    \renewcommand{\thetable}{A\arabic{table}}%
    \setcounter{figure}{0}
    \renewcommand{\thefigure}{A\arabic{figure}}%
    \setcounter{equation}{0}
    \renewcommand{\theequation}{A\arabic{equation}}
}

\clearpage
\beginsupplement
\appendix

\onecolumn
\section*{\Large Appendix}

In this appendix we provide:

\S~\ref{appendix:impl} additional implementation details of our method.

\S~\ref{appendix:baseline}: additional implementation details of the baseline methods.

\S~\ref{appendix:addi-quali}: additional visualization of the input and lifted scribbles.

\S~\ref{appendix:nerf}: our NeRF-based model results.    
    
\S~\ref{appendix:social}: discussion of the societal impact.

\section{Implementation details}
\label{appendix:impl}

\subsection{Multi-view image features $\bv_p^\text{MVS}$}
\label{appendix:mvs-net}
\begin{figure*}[b]
\centering
\includegraphics[width=0.94\linewidth]{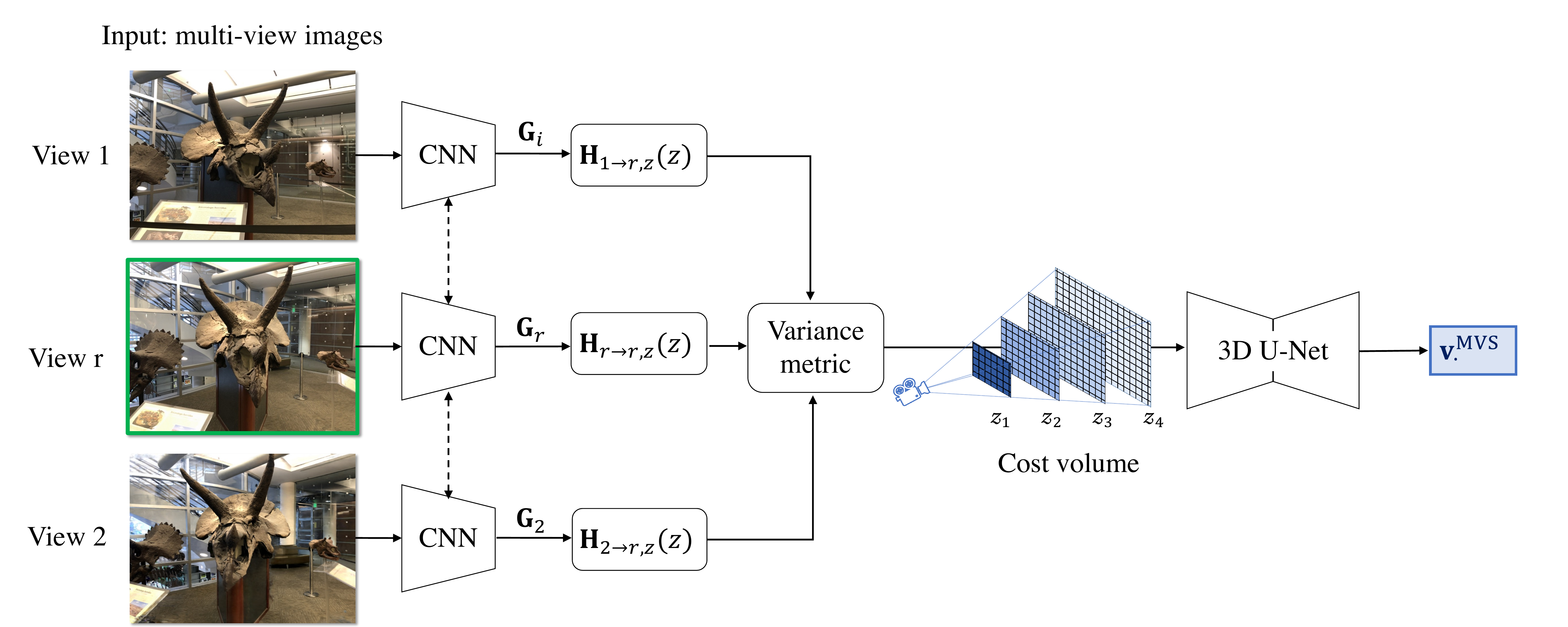}
\caption{\textbf{Process of computing multi-view image features $\bv_p^\text{MVS}$.}}
\label{fig:mvs-net}
\end{figure*}

We illustrate the process of computing multi-view image features $\bv_p^\text{MVS}$ in Fig.~\ref{fig:mvs-net}. We first extract 2D feature maps from multi-view images using a shared CNN model. Following~\cite{ChenICCV2021}, we use 3 close-by images as the multi-view set. Using more input images of different views is potentially beneficial, which we leave to future work.
We then project the computed 2D feature map into the reference view using Eq.~\eqref{eq:feat-warp} and compute the variance cost volume. To align with the MPI model used in this paper, we use the same 192 depth planes as suggested in NeX~\cite{Wizadwongsa2021NeX}.
Lastly, we extract the final features $\bv_p^\text{MVS}$ using a 3D U-Net. 
All the network architectures are detailed in Tab.~\ref{tab:net_arch}.

\subsection{Post-processing}
To ensure fast inference,  post-processing operates in a down-sampled and truncated volume space. We down-sample  the volume by $4\times$ (half the width and height) on the XY-plane using bi-linear interpolation. Based on the projected scribbles and the inferred 3D segmentation results, we first determine the null planes along the Z-axis where no foreground scribbles/predictions are located. Then for the remaining non-empty planes, we further down-sample to 20 planes using bi-linear interpolation over the Z-axis.  After post-processing, we sample the 3D volume back to its original resolution. The 3D graph-cut baseline is computed in the same way.

\begin{table}[t]
\centering
\small{
\begin{tabular}{cccccccc}
\specialrule{.15em}{.05em}{.05em}
\# & Layer & Kernel & Stride & Dilation & Input channels & Output channels & Input\\
\hline
\rowcolor{highlightRowColor} \multicolumn{8}{c}{\textit{Image encoder}} \\
\hline
1 & C2D/BN/ReLU & $3\times3$  & 1 & 1 & 3 & 8 & Image \\
2 & C2D/BN/ReLU & $3\times3$  & 1 & 1 & 8&8 & \#1 \\
3 & C2D/BN/ReLU & $5\times5$  & 2 & 2 & 8&16 & \#2 \\
4 & C2D/BN/ReLU & $3\times3$  & 1 & 1 & 16&16 & \#3 \\
5 & C2D/BN/ReLU & $3\times3$  & 1 & 1 & 16&16 & \#4 \\
6 & C2D/BN/ReLU & $5\times5$  & 2 & 2  & 16&32 & \#5 \\
7 & C2D/BN/ReLU & $3\times3$  & 1 & 1 & 32&32 & \#6 \\
8 & C2D  & $3\times3$  & 1 & 1 & 32&32 & \#7 \\
\hline
\rowcolor{highlightRowColor} \multicolumn{8}{c}{\textit{3D U-Net}} \\
\hline
9 & C3D/BN/ReLU & $3\times3$  & 1 & 1 & (32+9)&8 & \#8+3$\times$Images \\
10 & C3D/BN/ReLU & $3\times3$  & 2 & 1 & 8&16  & \#9 \\
11 & C3D/BN/ReLU & $3\times3$  & 1 & 1 & 16&16 & \#10 \\
12 & C3D/BN/ReLU & $3\times3$  & 2 & 1 & 16&32 & \#11 \\
13 & C3D/BN/ReLU & $3\times3$  & 1 & 1 & 32&32 & \#12 \\
14 & C3D/BN/ReLU & $3\times3$  & 2 & 1  & 32&64 & \#13 \\
15 & C3D/BN/ReLU & $3\times3$  & 1 & 1 & 64&64 & \#14 \\
16 & CT3D/BN/ReLU & $3\times3$  & 2 & 1 & 64&32 & \#15+\#14 \\
17 & CT3D/BN/ReLU & $3\times3$  & 2 & 1  & 32&16 & \#16+\#12 \\
18 & CT3D/BN/ReLU & $3\times3$  & 2 & 1 & 16&8 & \#17+\#10 \\
\specialrule{.15em}{.05em}{.05em}
\end{tabular}}
\caption{Network architectures. C2D is a 2D convolutional layer, C3D is a 3D convolutional layer, CT3D is a 3D de-convolutional layer, BN is a batch-normalization layer.}
\label{tab:net_arch}
\end{table}

\section{Baselines}
\label{appendix:baseline}

\paragraph{Graph-cut (2D).}
We use the unary term proposed in LazySnapping~\cite{li2004lazy}. Let the sets $\cF$ and $\cB$ denote the projected foreground and background scribbles. We first run K-means in the color space of these two pixel sets to get $K$ clusters. We refer to the cluster centers as $\bC_k^{\cB}\in\mathbb{R}^3$ and $\bC_k^{\cF}\in\mathbb{R}^3$ for cluster $k \in \{1, \dots, K\}$. Then, for each pixel $p$, we compute the minimum distance from its color $\bc_p\in\mathbb{R}^3$ to the foreground clusters as $d_p^{\cF} = \min_k\| \bc_p - \bC_k^{\cF}\|$; and similarly $d_p^{\cB} = \min_k\| \bc_p - \bC_k^{\cB}\|$. The unary term is then defined as:
\begin{equation}
\phi_p (y_p) =
\left\{
\begin{array}{ll}
 \phi_p (y_p=0) = \infty  & \forall p \in \cF\\
 \phi_p (y_p=1)= 0 & \forall p \in \cF\\
 \phi_p (y_p=0) = 0 & \forall p \in \cB\\
 \phi_p (y_p=1)= \infty & \forall p \in \cB\\
 \frac{(1-y_p)d_p^{\cB} + y_p d_p^{\cF} }{d_p^{\cB}+d_p^{\cF}} & \text{otherwise}\\
\end{array}
\right..
\label{eq:graphcut-2d-unary}
\end{equation}
%
For the binary term, we use the exponential term 
\begin{equation}
\psi_{p,q}(y_p,y_q) = |y_p - y_q| \cdot\exp^{-\frac{(\bc_p-\bc_q)^2}{\sigma}},
\label{eq:graphcut-2d-binary}
\end{equation}
where $\sigma$ is a balancing term which we set to 10. Practically, we find this exponential term outperforms the one proposed in the  LazySnapping~\cite{li2004lazy} formulation.

\paragraph{GrabCut~\cite{rother2004grabcut}.}
We use the GrabCut implementation provided in the OpenCV library.\footnote{\url{https://docs.opencv.org/3.4/d8/d83/tutorial\_py\_grabcut.html}} We provide the projected foreground and background scribbles as input, and use the mask segmentation mode (\texttt{GC\_INIT\_WITH\_MASK} in the OpenCV library) with 10 iterations. 

\paragraph{DeepLabV3~\cite{chen2017rethinking}.}
We use the COCO pre-trained model with ResNet-50 backbone provided in the  \texttt{torchvision} library.\footnote{\url{https://pytorch.org/vision/stable/models.html}} We fine-tune the network for binary classification using a binary-cross-entropy loss, where the positive and negative examples are  the projected foreground and background voxels respectively. We train the network for 20 epochs using Adam optimizer with a learning rate of  0.0001.

\paragraph{Deep Extreme Cut (DEXTR)~\cite{Man2018dextr}.}
We use the released model pre-trained on PASCAL VOC and SBD data. We compute the 4 extreme points from the projected foreground scribbles which are used as the network input.

\paragraph{FCA-Net~\cite{lin2020interactive}.} 
We randomly sample 100 points from the projected foreground and background scribbles respectively, and process them with a pre-trained FCA-Net (Res2Net~\cite{gao2019res2net} backbone). We run the baseline model 5 times and report the mean value.

\section{Additional visualization of the scribbles}
\label{appendix:addi-quali}
\begin{figure*}[t]
\centering
\includegraphics[width=0.94\linewidth]{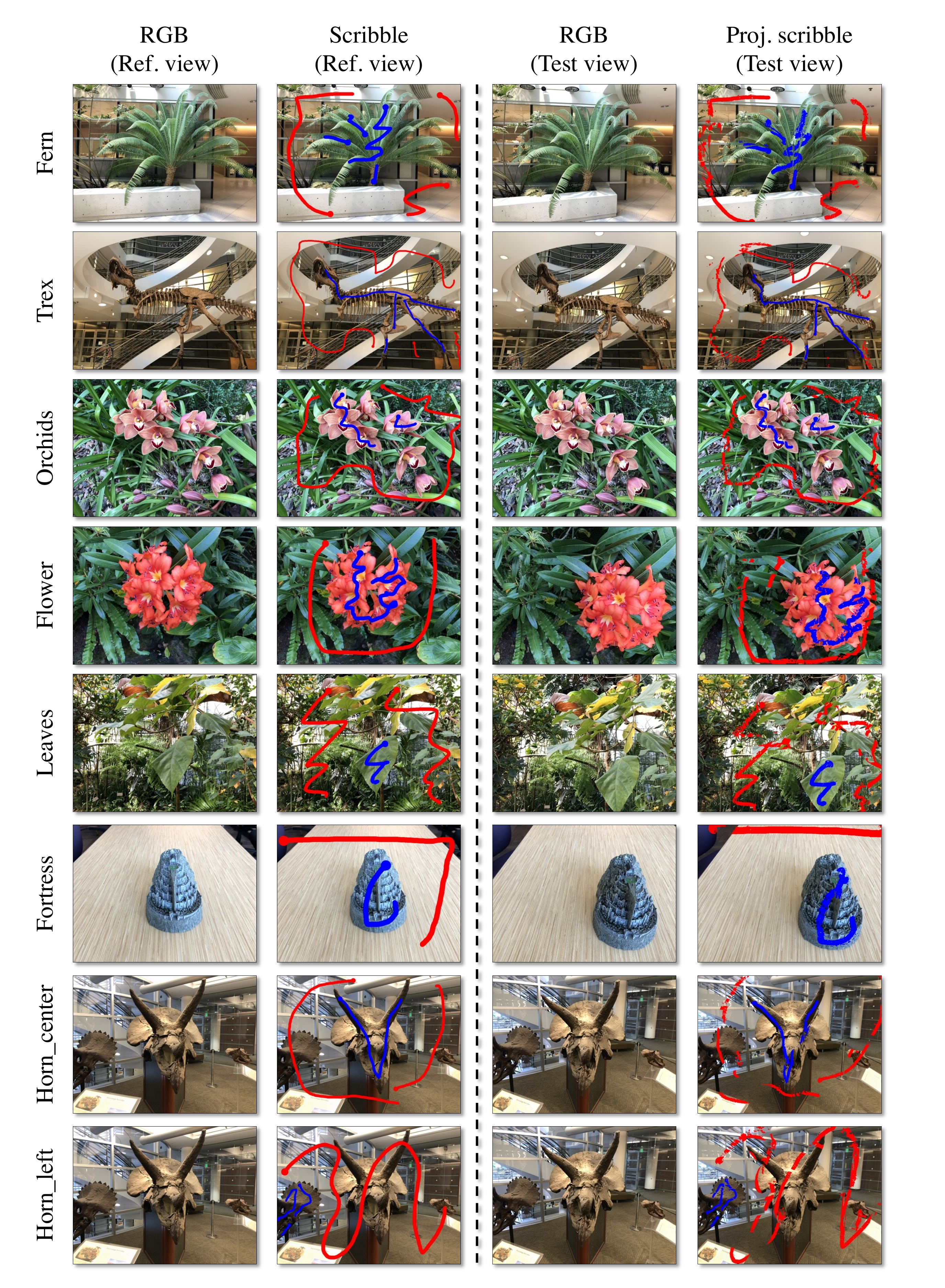}
\caption{Examples of the original and projected scribbles on the LLFF~\cite{mildenhall2019llff} dataset ({\color{blue} blue: foreground},   {\color{red} red: background}).}
\label{fig:scribble-llff}
\end{figure*}

For completeness, we illustrate the reference image and the input scribbles, together with the test-view image and the lifted scribbles in Fig.~\ref{fig:scribble-llff}.

\section{NeRF results}
\label{appendix:nerf}
\begin{figure*}[t]
\includegraphics[width=\linewidth]{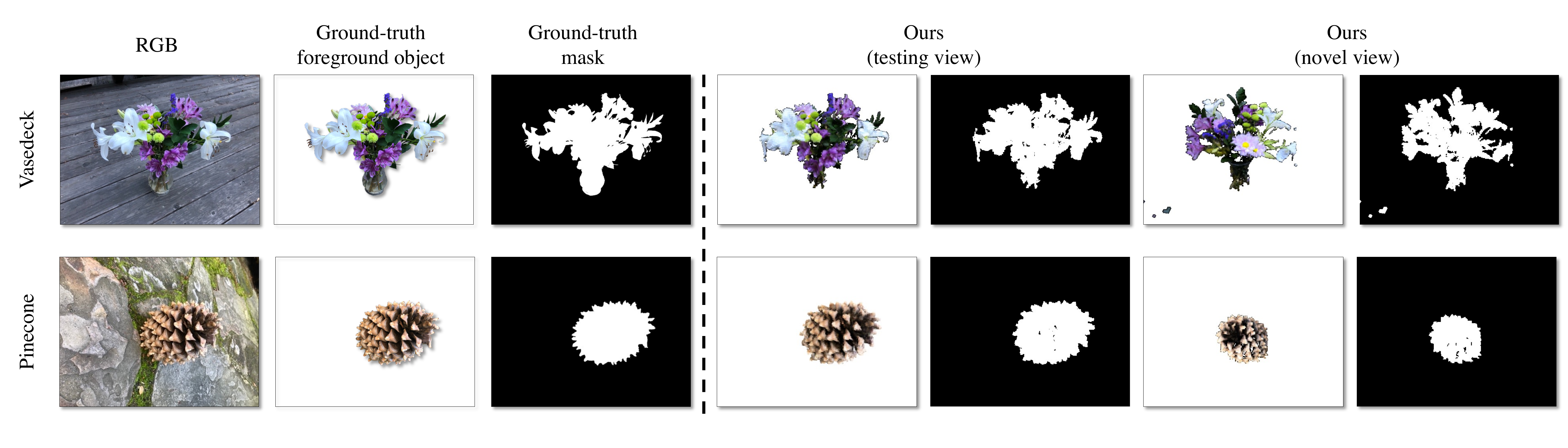}
\vspace{-0.3cm}
\caption{Qualitative results on the NeRF-real360~\cite{MildenhallECCV2020} dataset.}
\label{fig:res-nerf}
\end{figure*}

Our method also generalize to NeRF models. We use  PlenOctrees~\cite{yu2021plenoctrees} to extract the discrete volumetric representation from learned NeRF networks. For our purpose, we use the NeRF-real360 dataset which contains 2 real-world 360 scenes. Following PlenOctrees~\cite{yu2021plenoctrees}, we use the modified NeRF models where Spherical Harmonics are used to represent color rather than RGB values. We then convert the learned NeRF network into a $512^3$ volume following the suggested  PlenOctrees settings~\cite{yu2021plenoctrees}. 
We present the qualitative results in Fig.~\ref{fig:res-nerf} where our method correctly localizes and segments the foreground object in the scene.

\section{Societal impact}
\label{appendix:social}
In this work, we study novel-view object selection in neural volumetric representations. Our approach has the potential to positively impact applications in computer graphics and augmented reality, among them applying artistic effects on selected objects. However, our approach could also be used as a component for compositing objects into 3D scenes to create misinformation.

This research does not use human-derived data. The LLFF~\cite{mildenhall2019llff} dataset is released under the GNU General Public License v3.0. The Shiny~\cite{Wizadwongsa2021NeX} and NeRF-real 360~\cite{MildenhallECCV2020} datasets are released under MIT License. These datasets are mainly real-world scenes and do not contain offensive content or involve high-risk groups.

\end{document}